# Sub- Diving Labeling Method for Optimization Problem by Genetic Algorithm

**Masoumeh Vali**

Department of Mathematics, Dolatabad Branch, Islamic Azad University, Isfahan, Iran
E-mail: vali.masoumeh@gmail.com

**Abstract**

In many global Optimization Problems, it is required to evaluate a global point (min or max) in large space that calculation effort is very high. In this paper is presented new approach for optimization problem with subdivision labeling method (SLM) but in this method for higher dimensional has high computational. SLM Genetic Algorithm (SLMGA) in optimization problems is one of the solutions of this problem. In proposed algorithm the initial population is crossing points and subdividing in each step is according to mutation. RSLMGA is compared with other well known algorithms: DE, PGA, Grefensstette and Eshelman and numerical results show that RSLMGA achieve global optimal point with more decision by smaller generations.

**Keywords:** genetic algorithm (GA), Subdivision labeling method (SLM), crossing point.

## 1    Introduction

Global optimization has received a lot of attention in the past ten years, due to the success of new algorithms for solving large classes of problems from diverse areas such as computational chemistry and biology, structural optimization, computer sciences, operations research, economics, and engineering design and control.

In this paper, we are primarily concerned with the following optimization problem:
f ($x_1, x_2, ..., x_m$) where each $x_i$ is a real parameter object to $a_i \leq x_i \leq b_i$ for some constants $a_i$ and $b_i$.

Such problems have widespread application include optimization simulating models , fitting nonlinear curve to data , solving system of nonlinear , engineering design and control problem, and setting weights on neural networks. SLM method is new method for finding global minimum that performs more optimal than the other methods.

As compared to other optimization methods, Genetic Algorithm (GA) as an auto-adapted global searching system by simulating biological evolution and the fittest principle in natural environment seeks best solution more effectively [1]. Therefore, GA is a heuristic search technique that mimics the natural evolution process such as selection, crossover and mutation operations. The selection pressure drives the population toward better solutions while crossover uses genes of selected parents to produce offspring that will form the next generation. Mutation is used for avoiding of premature convergence and consequently escaping from local optimal. The GAs have been very successful in handling combinatorial optimization problems which are difficult [2]. SLM method in high dimension have high search space, GA provides a comprehensive search methodology for optimization.

GA is applicable to continuous optimization problems. In global optimization scenarios, GAs often manifests their strengths: efficient, parallelizable search; the ability to evolve solutions with different dimension; and a characterized and controllable process of innovation.

This paper starts with the description of related work in section 2. Section 3 gives the outline of Model and Problem Definition SLM and text problems of SLM method and comparision of SLM method with RS, RSW and SA [3, 4]. In section 4, we have a discussion about how genetic algorithm can be used to SLM. In section 5, we present schemata for SLM with GA. In section 6, test functions used in simulation studies and test problems of De Jong by SLMGA method are in sections 6 and 7. In section 8 is shown Experimental Results and comparison of SLMGA with other methods: DE, PGA, Grefensstette and Eshelman [5]**.**The discussion ends with a conclusion and future trend.

## 2  Related work

In the early 1960s and 1970s, new search algorithms were initially proposed by Holland, his colleagues and his students at the University of Michigan. These search algorithms which are based on nature and mimic the mechanism of natural selection were known as Genetic Algorithms (GAs) [1, 3, 6, 7, 8]. Holland in his book "Adaptation in Natural and Artificial Systems" [1] initiated this area of study. Theoretical foundations besides exploring applications were also presented.

The population of solutions is initialized by applying the GAs operators such as the crossover and mutation [1, 3, 6]. And with their natural selection they have an iterative procedure usually used to optimize and select the best chromosome (solution) in the population. This population consists of various solutions to hard complex problems and is usually generated randomly [6, 9].

Figure (1) below represents the Simple Standard GA evolution flow.

Zhang et al.[10] introduced triangulation theory into GA by virtue of the concept of relative coordinate genetic coding , design corresponding crossover and mutation operator.

Hayes and Gedeon [11] considered infinite population model for GA where the generation of the algorithm corresponds to a generation of a map. They showed that for a typical mixing operator all the fixed points are hyperbolic.

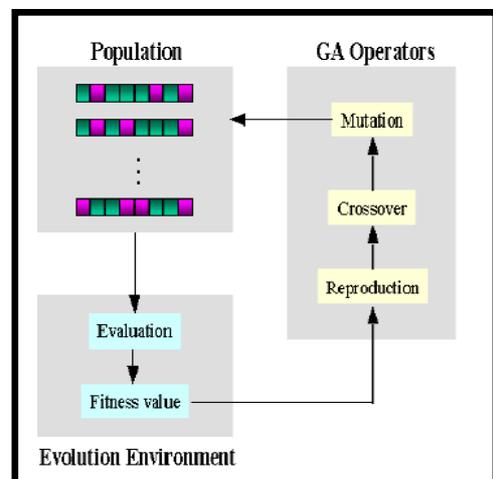

Figure 1: Evolution flow of genetic algorithm [5].

Gedeon et al. [12] showed that for an arbitrary selection mechanism and a typical mixing operator, their composition has finitely many fixed points.

Qian et al.[13] proposed a GA to treat with such constrained integer programming problem for the sake of efficiency. Then, the fixed-point evolved (E)-UTRA PRACH detector was presented, which further underlines the feasibility and convenience of applying this methodology to practice.

## 3   Model and Problem Definition SLM

In this section, Model of SLM is expressed then SLM is implemented on the Goldstein-Price function and the Easom function and be traced all operations step by step. At the end, SLM method is compared three methods: RS, RSW [3] and SA and results are shown.

At present, Supposing F($x_1, x_2, x_3 \ldots, x_n$) with constraint $a_i \leq x_i \leq b_i$, we want to earn global point to this in order to following serial algorithm SLM:

**Step1:** Draw the diagrams for $x_i = b_i$ and $x_i = a_i$ for i=1, 2,…, n ; then, we find crossing points which equal $n^2$ and h= min$\frac{|x_i| + |x_j|}{2}$ for $1 \leq i, j \leq n$.

**Step2:** Suppose that the point$(a_1, b_1, b_2 \ldots b_{n-1})$ is one of the crossing points. Consider the values of $\pm h$ gained by step1 and then do the algebra operations on this crossing point as shown in Table 1. Their maximum number in dimensional space is $2 * (\binom{n}{1} + \binom{n}{2} + \binom{n}{3} + \cdots + \binom{n}{n}))$.

Table 1: The number of point is produced by algebra

| |
|---|
| $(a_1 \pm h, b_1, b_2 \ldots b_{n-1})$ |
| $(a_1, b_1 \pm h, b_2 \ldots b_{n-1})$ |
| ⋮ |
| $(a_1, b_1, b_2 \ldots b_{n-1} \pm h)$ |
| ⋮ |
| $(a_1 \pm h, b_1 \pm h, b_2 \ldots b_{n-1})$ |
| ⋮ |
| $(a_1 \pm h, b_1 \pm h, b_2 \pm h, \ldots, b_{n-1} \pm h)$ |

**Step 3:** The function value is calculated for all points of step 2 and the value of these functions is compared with f$(a_1, b_1, b_2 \ldots b_{n-1})$. At the end, we will select the point which has the minimum value and we will call it$(c_1, c_2, c_3 \ldots c_n)$.

**Note:** If we want to find the optimal global max, we should select the maximum value.

**Step 4:** In this step, the equation 1 is calculated:

$$(c_1, c_2, c_3 \ldots c_n) - (a_1, b_1, b_2 \ldots b_{n-1}) = (d_1, d_2, d_3 \ldots d_n) \quad (1)$$

**Step 5:** According to the result of step 4, the point $(a_1, b_1, b_2 \ldots b_{n-1})$ is labeled according to equation (2).

$$l(a_1, b_1, b_2 \ldots b_{n-1}) = \begin{cases} 0 & d_1 \geq 0, \ldots, d_n \geq 0 \\ 1 & d_1 < 0, d_2 \geq 0, \ldots, d_n \geq 0 \\ 2 & d_2 < 0, d_3 \geq 0, \ldots, d_n \geq 0 \\ \vdots & \\ N & d_n < 0 \end{cases} \quad (2)$$

**Step 6:** Go to step 7 if all the crossing points were labeled, otherwise repeat the steps 2 to 5.

**Step 7:** In this step, the complete labeling polytope is focused. In fact, a polytope will be chosen that has complete labeling in different dimensions as shown in Table 2.

Table 2: complete label with different dimensional

| Dimension | Complete Label |
|---|---|
| 2 | 0,1,2 |
| 3 | 0,1,2,3, |
|  |  |
| n | 0,1,2,3,…,n |

**Step 8:** In this step, all sides of the selected polytope (from step 7) are divided into 2 according to equation 3 and we repeat steps 3 to 7 for new crossing points.

$$h = \min\left\{\frac{|x_i| + |x_j|}{2}\right\} \quad 1 \leq i, j \leq n \quad (3)$$

**Step 9:** Steps 2 to 8 are repeated to the extent that h → 0 and the result is global min or max point.

## 3.1 Test Problems of SLM:

In this section, we have three test problems that is implemented by SLM.

### 3.1.1 Test Problem 1

The Goldstein-Price function [GP71] is a global optimization test function.
function definition:
$$f_{\text{Gold}}(x_1, x_2) = \left(1 + (x_1 + x_2 + 1)^2 . (19 - 14x_1 + 3x_1^2 - 14x_2 + 6x_1x_2 + 3x_2^2)\right).$$

$$\left(30 + (2x_1 - 3x_2)^2 . (18 - 32x_1 + 12x_1^2 + 48x_2 - 36x_1x_2 + 27x_2^2)\right)$$

Global minimum:
$$f(x_1, x_2) = 3; \quad (x_1, x_2) = (0, -1)$$

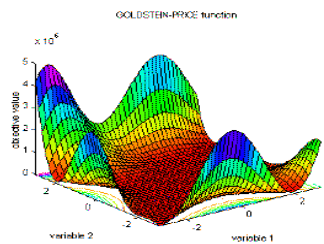

Figure 2: Figure of the Goldstein-Price

| | Table 3: Initial Population of $f_{\text{Easo}}$ | | | | |
|---|---|---|---|---|---|
| | $h_1=4$, P(0): | $h_2=2$ | P(1): | $l(x)$ | Fixed Point |
| | (-2,-2) | $\xrightarrow{M}$ | (0,-2) | 0 | |
| | (2,2) | | (0,0) | 2 | (0,0) |
| | (-2,2) | | (0,0) | 2 | |
| | (2,-2) | | (0,-2) | 1 | |
| Figure 3: Initial Population of $f_{\text{Easo}}$ | | | | | |

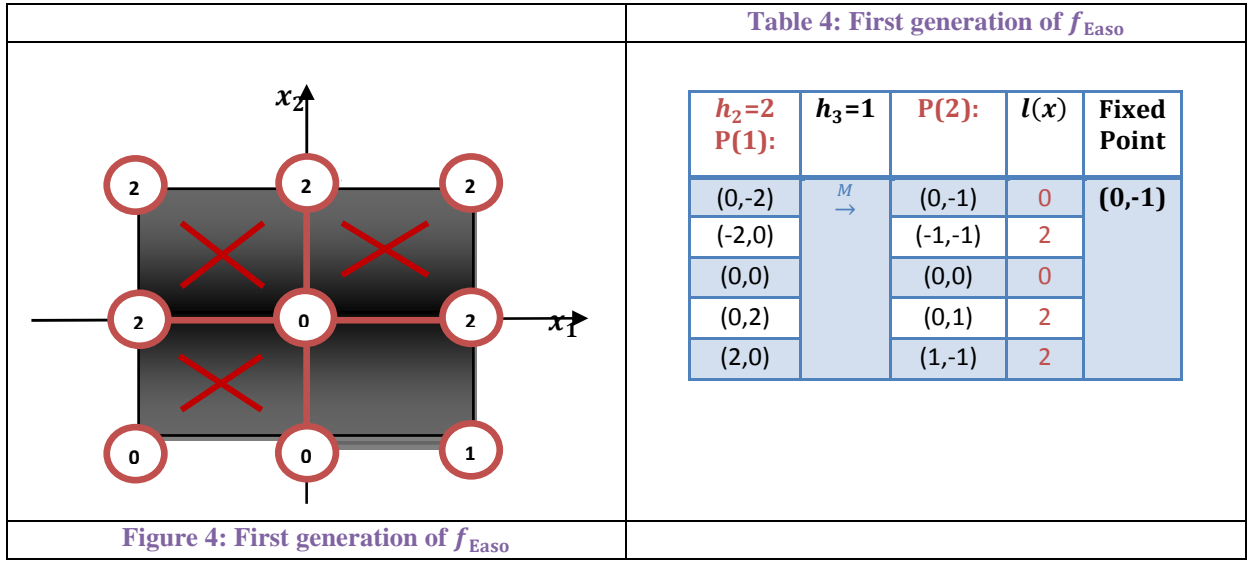

Figure 4: First generation of $f_{Easo}$

Table 4: First generation of $f_{Easo}$

| $h_2=2$ P(1): | $h_3=1$ | P(2): | $l(x)$ | Fixed Point |
|---|---|---|---|---|
| (0,-2) | $\xrightarrow{M}$ | (0,-1) | 0 | **(0,-1)** |
| (-2,0) | | (-1,-1) | 2 | |
| (0,0) | | (0,0) | 0 | |
| (0,2) | | (0,1) | 2 | |
| (2,0) | | (1,-1) | 2 | |

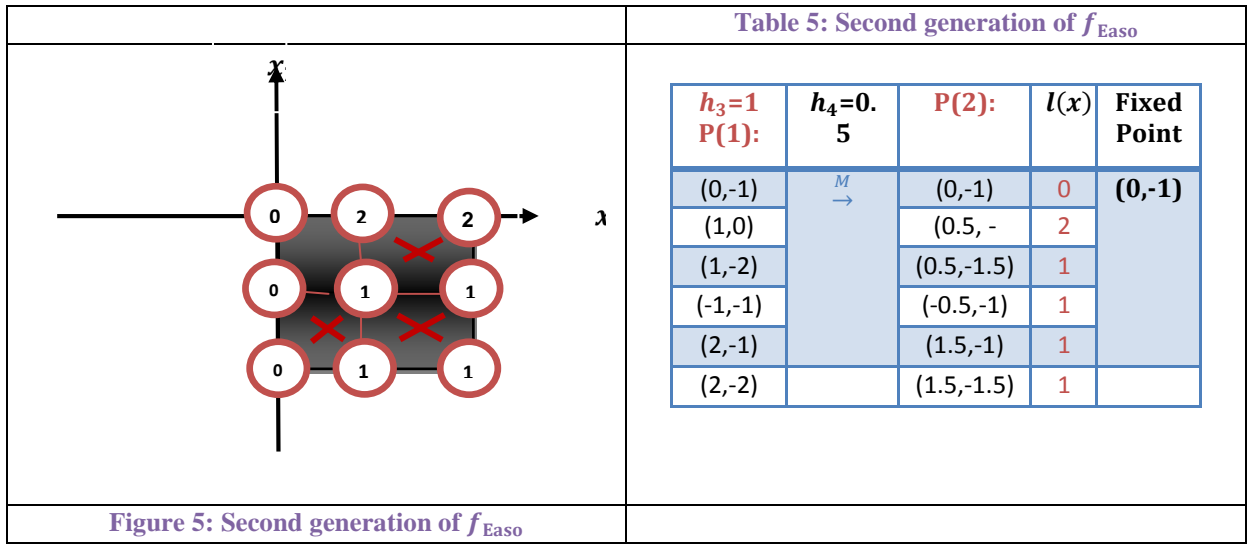

Figure 5: Second generation of $f_{Easo}$

Table 5: Second generation of $f_{Easo}$

| $h_3=1$ P(1): | $h_4=0.5$ | P(2): | $l(x)$ | Fixed Point |
|---|---|---|---|---|
| (0,-1) | $\xrightarrow{M}$ | (0,-1) | 0 | **(0,-1)** |
| (1,0) | | (0.5, -) | 2 | |
| (1,-2) | | (0.5,-1.5) | 1 | |
| (-1,-1) | | (-0.5,-1) | 1 | |
| (2,-1) | | (1.5,-1) | 1 | |
| (2,-2) | | (1.5,-1.5) | 1 | |

Table 6: comparison test problem #1 among three methods

| Algorithms | Iteration | Optimal point | Best Solution | Standard deviation |
|---|---|---|---|---|
| SLM | 17 | (-0.99993896484, 0.00006103516) | | (0.00006103516, 0.00006103516) |
| RS | 1000 | (-1.564, 0.999897) | | (0.564, 0.000103) |
| $RSW(x^{initial} = (14.0356, 14.0356))$ | 500 | (-1, 0.099879) | (-1,0) | (0, 0.099879) |
| SA | 400 | (-0.99997948942, 0.00003051758) | | (0.00003051758, 0.00003051758) |

### 3.1.2 Test problem2

The Easom function [Eas90] is a unimodal test function, where the global minimum has a small area relative to the search space. The function was inverted for minimization.

function definition:

$$f_{Easo}(x_1, x_2) = -\cos(x_1).\cos(x_2).e^{-((x_1-\pi)^2+(x_2-\pi)^2)}; \qquad -100 \leq x_i \leq 100, i = 1,2$$

global minimum:

$$f(x_1, x_2) = -1; \quad (x_1, x_2) = (\pi_i, \pi_i)$$

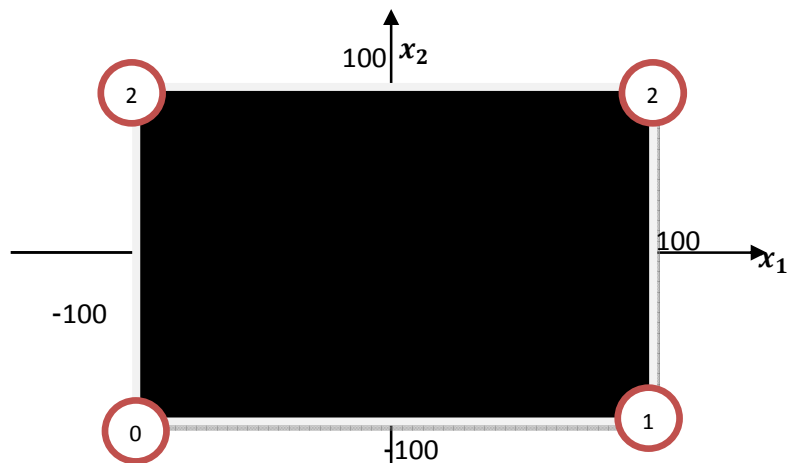

Figure 6 : Initial Population of $f_{Easo}$

Table 7: Initial Population of $f_{Easo}$

| $h_1$=200, P(0): | $h_2$=100 | P(1): | $l(x)$ | Solution |
|---|---|---|---|---|
| (-100,-100) | $\xrightarrow{M}$ | (0,0) | 0 | **(0,0)** |
| (100,100) |  | (0,0) | 2 |  |
| (-100,100) |  | (0,0) | 2 |  |
| (100,-100) |  | (0,0) | 1 |  |

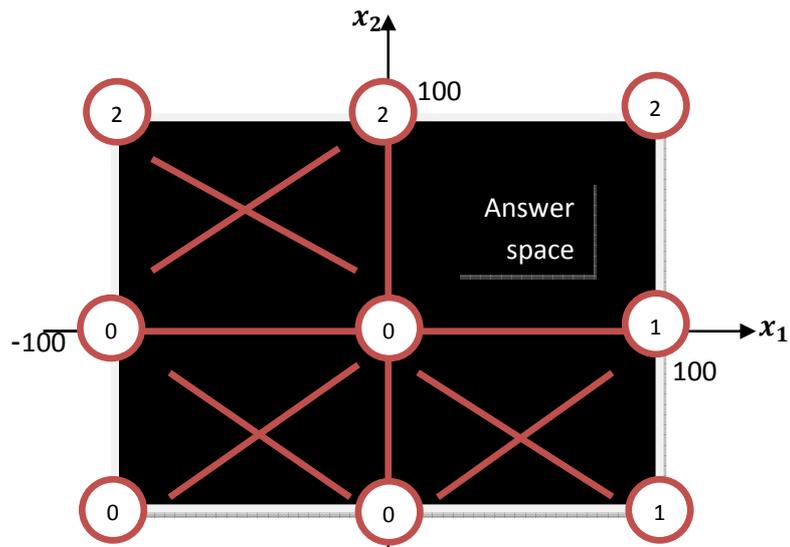

Figure 7: First generation of $f_{Easo}$

Table 8: First generation of $f_{Easo}$

| $h_2=100$ P(1): | $h_3=50$ | P(2): | $l(x)$ | Solution |
|---|---|---|---|---|
| (-100,0) | $\xrightarrow{M}$ | (-50,0) | 0 | (0,0) |
| (0,-100) |  | (0,-50) | 0 |  |
| (100,0) |  | (50,0) | 1 |  |
| (0,100) |  | (0,50) | 2 |  |
| (0,0) |  | (0,0) | 0 |  |

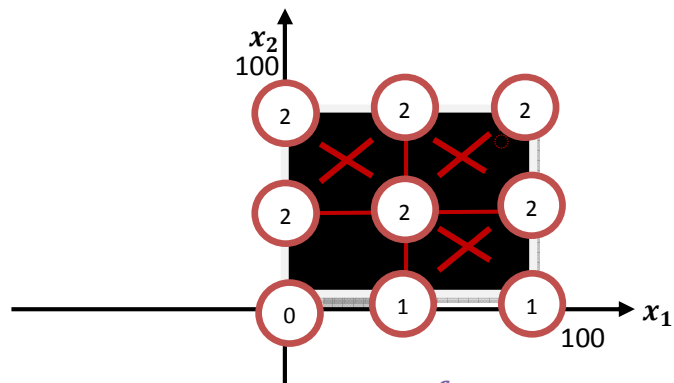

Figure 8: Second generation of $f_{Easo}$

Table 9: First generation of $f_{Easo}$

| $h_3=50$, P(2): | $h_4=25$ | P(3): | $l(x)$ | Solution |
|---|---|---|---|---|
| (0,50) | $\xrightarrow{M}$ | (0,25) | 2 | (0,0) |
| (50,0) |  | (25,0) | 1 |  |
| (50,50) |  | (25,25) | 2 |  |
| (50,100) |  | (25,75) | 2 |  |
| (100,50) |  | (75,25) | 2 |  |

Table 10: comparison test problem# 2 among three methods

| Algorithms | Iteration | Optimal point | Best point | Standard deviation |
|---|---|---|---|---|
| SLM | 11 | (3.3203125,3.3203125) | $(\pi_i, \pi_i)$ | (0.1803125,0.1803125) |
| RS | 1500 | (3.99987, 3.9876988) | | (0.85987, 0.8476988) |
| RSW($x^{initial}=(1.048, 0.89)$) | 700 | (3,3) | | (.014,0.14) |
| SA | 1200 | (3,3) | | (.014,0.14) |

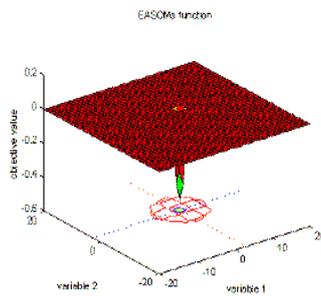

Figure 9: Figure of the Easom function

## 4 Schema of SLMGA:

### 4.1 Initialization

Initially many individual solutions are generated to form an initial population. The population size depends on the nature of the problem; the initial population in SLMGA is generated according to dimension for example initial population for 2- dimensional is
 four, for 3- dimensional is eight and in this manner for n- dimensional is $2^n$ according to Table 11.Initial population in this method earns through crossing points with Table 11. In genetic algorithm like other evolutionary algorithms, its optimal solutions are points that the algorithm improves keeps or returns to them after a certain number of iterations because these points meet required criteria of the algorithm.

Table 11: The initial population in SGA

| Dimensional | Crossing points |
|---|---|
| 2 | $2^2$ |
| 3 | $2^3$ |
| 4 | $2^4$ |
| ⋮ | ⋮ |
| n | $2^n$ |

## 4.2 Selection

During each successive generation, a proportion of the existing population is selected to breed a new generation. Individual solutions are selected through a fitness-based process, where fitter solutions (as measured by a fitness function) are typically more likely to be selected. Certain selection methods rate the fitness of each solution and preferentially select the best solutions.

Supposing that algorithm is searching a point x which can make continuous function of $f$ to achieve its minimum. The necessary and sufficient condition of extreme point is that this point gradient is 0 that is $\nabla f(x) = 0$.

For self-mapping $g: \mathbb{R}^n \to \mathbb{R}^n$, we say $x \in \mathbb{R}^n$ is a fixed point of $g$ if $g(x) = x$, then we can convert the solution of zero point problems to fixed point ones of function $g(x) = x + \nabla f(x)$.

Supposing that definition domain of $f(x_1, x_2, \ldots, x_n)$ is that $a_1 \leq x_1 \leq a_2$, $a_3 \leq x_2 \leq a_4, \ldots, a_r \leq x_n \leq a_s$ and dividing the domain into many polytopes with two groups of straight lines of $\{x_1 = mh_i\}, \{x_2 = mh_i\}, \ldots, \{x_n = mh_i\}$ in which m is a not negative integer and $h_i$ is a positive quantity relating to precision of the problem. As a result, we can code each point of intersection as $x_1 = a_1 + k_1 h_i, x_2 = a_3 + k_2 h_i, \ldots, x_n = a_r + k_n h_i$ where $k_1, k_2, \ldots, k_n$ are not negative integers, so $(k_1, k_2, \ldots, k_n)$ is called the relative coordinates of points. Consequently, by changing $k_1, k_2, \ldots, k_n$ relative coordinates of each point in search space is determined.

Attentive points that are pointed above, labeling is done according equation (2):

$$l(x) = \begin{cases} 0, & g_1(x)-x_1 \geq 0, \ldots, g_n(x)-x_n \geq 0 \\ 1, & g_1(x)-x_1 < 0, g_2(x)-x_2 \geq 0, \ldots, g_n(x)-x_n \geq 0 \\ 2, & g_2(x)-x_2 < 0, g_3(x)-x_3 \geq 0, \ldots, g_n(x)-x_n \geq 0 \\ \vdots & \\ n, & g_n(x)-x_n < 0 \end{cases} \quad (2)$$

The square with all different kinds of integer label is called a completely labeled unite, when $h_i \to 0$ within iteration stages, vertices of that square approximately converge to one point which is a fixed point.

### 4.3 Mutation Operator

For each point coded $(k_1, k_2, \ldots, k_n)$, the GA is trying to improve it to reach optimal solution by mutation operator searching all points surrounding it in certain step determined by $h_{i+1}$.

For instance, $(k_1, k_2, \ldots, k_n)$ in P(0), initial population, addressing $(x_1 + k_1 h_i, x_2 + k_2 h_i, \ldots, x_n + k_n h_i)$ will be changed as $(x_1 + \alpha_1, x_2 + \alpha_2, \ldots, x_n + \alpha_n), \alpha_1, \alpha_2, \ldots, \alpha_n \in \{0, \pm h_{i+1}\}$. Subsequently, the algorithm saves the best-mutated individual among all possible offspring. Therefore, this operator produces new population located on intersection of the next grid. Because of this, coming polytopes are specified to evaluate and label. Furthermore, the next generation is producing from the previous one.

## 5  Schemata Analysis for SLMGA

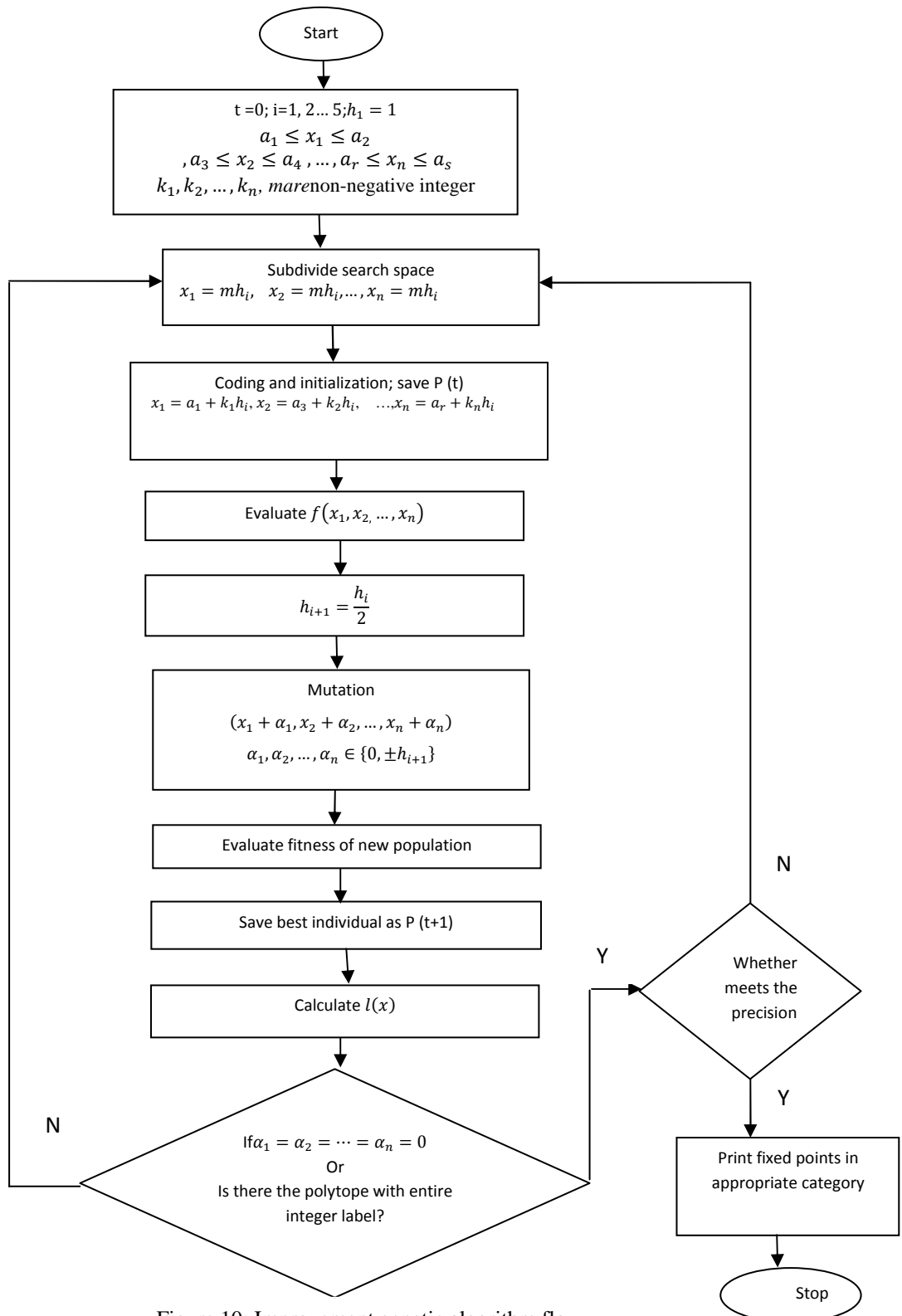

Figure 10: Improvement genetic algorithm flow

This improved algorithm makes grid in given scope and encodes each intersection by integer while it starts from the lowest point of the domain. After calculating fitness of each point, it generates the best offspring and computes integer label of the last population for every polytope. When it found the polytope labeled completely, subdivides them in order to seek the solution closely. As following, we demonstrate the performance of the improved algorithm by different examples and show how it can categorize fixed points. This schema is shown in Figure 10.

## 6 Test Function Used in Simulation Studies

Five test functions (F1-F5) presented in Table 12 have been firstly proposed by De Jong [15] to measure the performance of GAs. After Jong, they have been extensively used by GA researchers and other algorithm communities. The test environment includes functions which are convex (F1), non-convex (F2), discontinuous (F3), stochastic (F4) and multimodal (F5).

Table12. The structure of the first three Dejong

| Function Number | Function | Limits | Dim. | Initial Population |
|---|---|---|---|---|
| F1 | $\sum_{i=1}^{3} x_i^2$ | $-5.12 \leq x_i \leq 5.12$ | 3 | 8 |
| F2 | $100.(x_1^2 - x_2) + (1 - x_1)^2$ | $-2.048 \leq x_i \leq 2.048$ | 2 | 4 |
| F3 | $30. + \sum_{i=1}^{5} \lfloor x_j \rfloor$ | $-65.536 \leq x_i \leq 65.536$ | 5 | 32 |
| F4 | $\sum_{i=1}^{30} (ix_i^4. + Gauss(0,1))$ | $-1.28 \leq x_i \leq 1.28$ | 30 | |
| F5 | $\dfrac{1}{0.002 + \sum_{i=0}^{24} \frac{1}{i + \sum_{j=0}^{1}(x_j - a_{ij})^6}}$ | $-65.536 \leq x_i \leq 65.536$ | 2 | 4 |

## 7 Implementing SLMGA on De Jong's functions:

In this section, SLMGA is implemented on F1 and F5.

### 7.1 Test problem 1

Equation (3) is De Jong's second function (Rosenbrock's saddle) that by SLM is implemented.

(3)

$$f_2(x_1, x_2) = 100.(x_1^2 - x_2) + (1 - x_1)^2; \quad x_j \in [-2.048, 2.048]$$

Although $f_2(x_1, x_2)$ has just two parameters, it has the reputation of being a difficult minimization problem. The minimum is $f_2(1,1) = 0$. In this function; figures from 11 to 14 are shown.

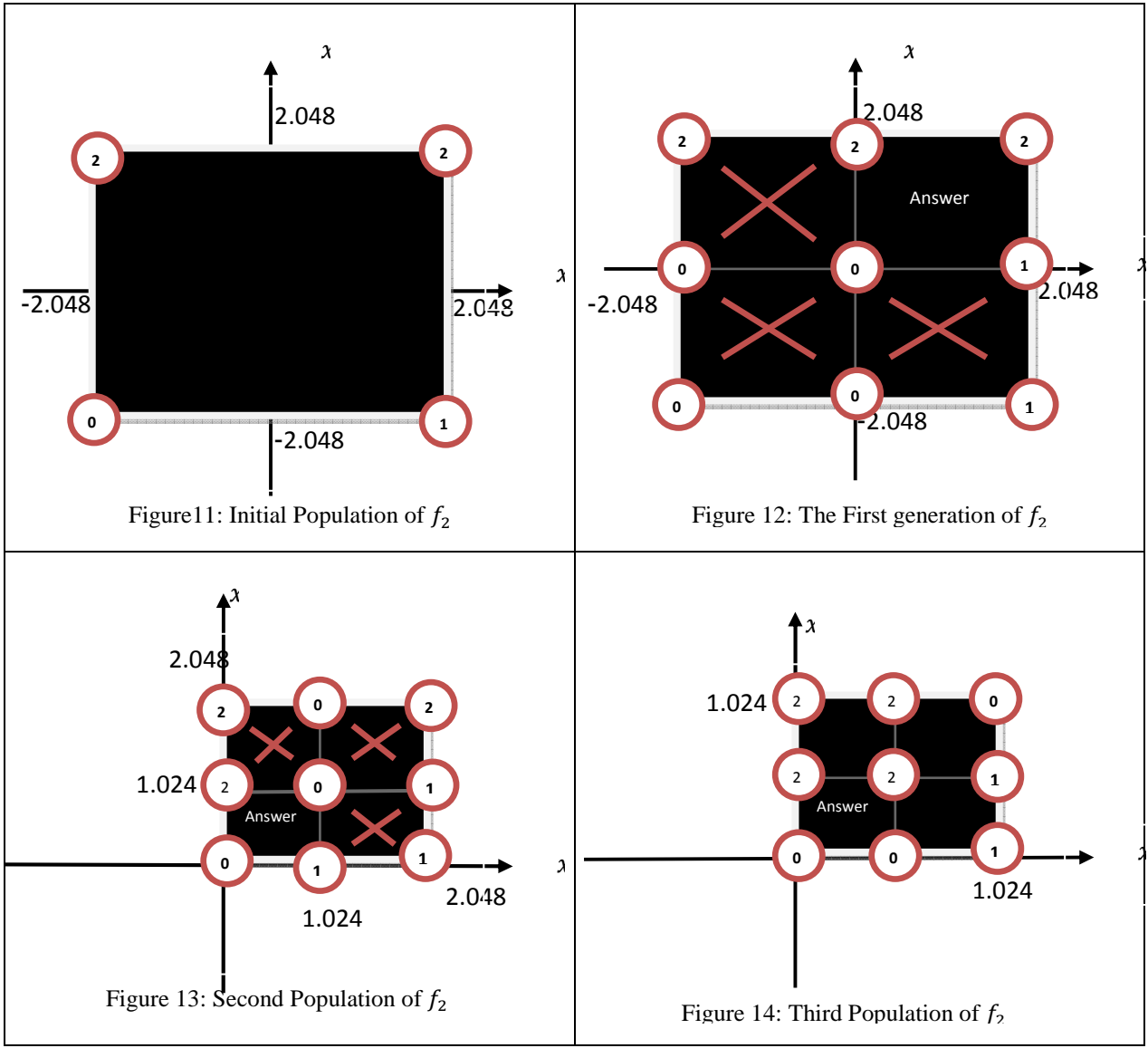

Figure 11: Initial Population of $f_2$

Figure 12: The First generation of $f_2$

Figure 13: Second Population of $f_2$

Figure 14: Third Population of $f_2$

### 7.2 Test problem 2

This is test problem of Fifth De Jong function (Shekel's Foxholes).

$$f_5(x_0, x_1) = \frac{1}{0.002 + \sum_{i=0}^{24} \frac{1}{i + \sum_{j=0}^{1}(x_j - a_{ij})^6}}; \quad x_j \in [-65.536, 65.536] \quad (4)$$

With $a_{i0} = \{-32, -16, 0, 16, 32\}$ for $i = 0,1,2,3,4$ and $a_{i0} = a_{i \bmod 5, 0}$

As well as $a_{i1} = \{-32, -16, 0, 16, 32\}$ for $i = 0,5,10,15,20$ and $a_{i1} = a_{i+k,1}, k = 1,2,3,4$

The global minimum for this function is $f_5(-32, -32) \cong 0.998004$.

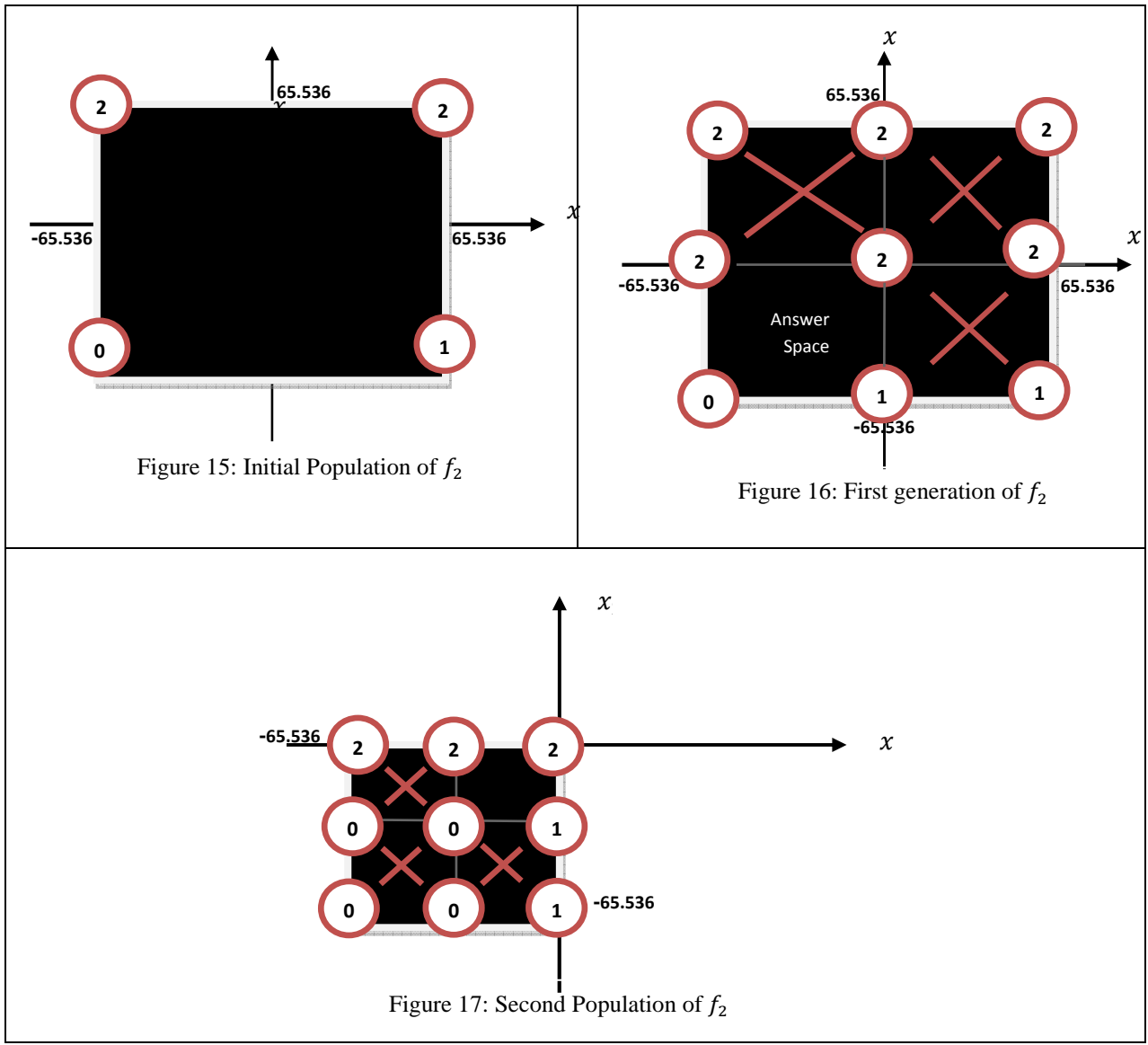

Figure 15: Initial Population of $f_2$

Figure 16: First generation of $f_2$

Figure 17: Second Population of $f_2$

## 8    Evolution

In this section, we present the experimental results of the proposed approach for solving some nonlinear numerical functions. Also, this approach was implemented by modifying the Genesis program of John Grefenstette.
In this implementing, all problems (F1 to F5) [De Jong, 1975] were run with the elitist strategy, complete label's selection procedure and a population size depend on dimensional space according Table 12. As table 13 shows, for all functions, each experiment consisted of running the genetic algorithm for 100 trials (function evaluations). Results were saved for the best performance; the best performance (BV) is the smallest value of the objective function obtained over all function evaluations. In order to select good parameter settings, multi runs were done for the SLMGA with different settings for mutation sizes (real) but mutation rates in all of experimental is 0.5 because our mutation became half than before step every time.

In last, standard deviation (SD) is calculated and measured with final answer of De Jong function.

Table13: Numerical result of De Jongs' function by SLMGA

| Step | Algorithm | De Jong's Function | Mutation Size | Mutation Rate | Best Point | BV | SD |
|---|---|---|---|---|---|---|---|
| Step1 | SLMGA | F1 | 10.24 | 0.5 | (0,0,0) | - | - |
| Step2 | | | 5.12 | | (0,0,0) | - | - |
| Step3 | | | 2.56 | | (0,0,0) | - | - |
| Step18 | | | 0.000078125 | | (0,0,0) | 0 | 0 |
| Step1 | SLMGA | F2 | 4.096 | 0.5 | (0,0) | - | - |
| Step2 | | | 2.048 | | (1.024,1.024) | - | - |
| Step3 | | | 1.024 | | (1.024,1.024) | - | - |
| | | | | | | | - |
| Step8 | | | 0.032 | | | 0 | 0.0063140 1 |
| Step1 | SLMGA | F3 | 10.24 | 0.5 | (-5.12,-5.12,-5.12, -5.12, -5.12) | - | - |
| Step2 | | | 5.12 | | (-5.12,-5.12,-5.12, -5.12, -5.12) | - | - |
| Step3 | | | 2.56 | | (-5.12,-5.12,-5.12, -5.12, -5.12) | - | - |
| Step10 | | | 0.02 | | (-5.10,-5.10,-5.12, -5.10, -5.10) | 0 | 0 |
| Step1 | SLMGA | F4 | 2.56 | 0.5 | $\underbrace{(0,0,\ldots,0)}_{30}$ | - | - |
| Step2 | | | 1.28 | | $\underbrace{(0,0,\ldots,0)}_{30}$ | - | - |
| Step3 | | | 0.64 | | $\underbrace{(0,0,\ldots,0)}_{30}$ | - | - |
| | | | | | | - | - |
| Step16 | | | | | $\underbrace{(0,0,\ldots,0)}_{30}$ | Depend on $\eta$ | 0 |
| Step1 | SLMGA | F5 | 131.072 | 0.5 | (0,0) | - | - |
| Step2 | | | 65.536 | | (-32.768,-32.768) | - | - |
| Step3 | | | 32.768 | | (-32.768,-32.768) | - | - |
| | | | | | | - | - |
| Step9 | | | 0.512 | | (-32.256,-32.256) | 1.000024 | 0.002236 |

We have compared the performance of the RSLMGA to that of other some well known genetic algorithms: DE, PGA, Grefensstette and Eshelman [5]. PGA, Grefensstette and Eshelman algorithms were run 50 times; the DE algorithm was run 1000 times and SLMGA 50 times for each function to achieve average results[5].

When the results produced by the RSLMGA for all functions were evaluated together, it was observed that the best of the average of number of generations for De Jong's functions.

It is seen in Table 14 that the convergence speed of the SLMGA achieves the optimal point with more decision by smaller generation.

As Table 14 the most significant improvement is with F4, proportion of the number of generations (PNG) about $\frac{2300}{16} \cong 144$ times smaller average of number of generations than the DE algorithm.

Table14: Comparison of Number of iterations among SLMGA and other methods

| Algorithms | F1 | F2 | F3 | F4 | F5 |
|---|---|---|---|---|---|
| $PGA(\lambda = 4)$ | 1170 | 1235 | 3481 | 3194 | 1256 |
| $PGA(\lambda = 8)$ | 1526 | 1671 | 3634 | 5243 | 2076 |
| Grefensstette | 2210 | 14229 | 2259 | 3070 | 4334 |
| Eshelman | 1538 | 9477 | 1740 | 4137 | 3004 |
| DE(F: RandomValues) | 260 | 670 | 125 | 2300 | 1200 |
| SLMGA | 18 | 8 | 10 | 16 | 9 |
| PNG | 15 | 84 | 13 | 144 | 134 |

**Conclusion**

SLMGA is a new method for finding the true optimal global optimization is based on Subdividing Labeling Method (SLM). In this work, the performance of the SLMGA has been compared to that of some other well known GAs. From the simulation studies, it was observed that SLMGA achieve the optimal point with more decision by smaller generation. Therefore, SLMGA seems to be a promising approach for engineering optimization problems.